\documentclass[10pt,twocolumn,letterpaper]{article}

\usepackage{iccv}              % To produce the CAMERA-READY version

\usepackage{amsthm}

\usepackage{multirow}
\usepackage{bm}

\newtheorem{theorem}{Theorem}

\newtheorem{definition}{Definition}

\usepackage{makecell}

\definecolor{iccvblue}{rgb}{0.21,0.49,0.74}
\usepackage[pagebackref,breaklinks,colorlinks,allcolors=iccvblue]{hyperref}

\def\paperID{7044} % *** Enter the Paper ID here
\def\confName{ICCV}
\def\confYear{2025}

\title{Membership Inference Attacks with False Discovery Rate Control}

\author{Chenxu Zhao \qquad Wei Qian \qquad Aobo Chen \qquad Mengdi Huai\\
Department of Computer Science \\
Iowa State University\\
{\tt\small \{cxzhao, wqi, aobochen, mdhuai\}@iastate.edu}
}

\begin{document}
\maketitle
\begin{abstract}

Recent studies have shown that deep learning models are vulnerable to membership inference attacks (MIAs), which aim to infer whether a data record was used to train a target model or not. To analyze and study these vulnerabilities, various MIA methods have been proposed. Despite the significance and popularity of MIAs, existing works on MIAs are limited in providing guarantees on the false discovery rate (FDR), which refers to the expected proportion of false discoveries among the identified positive discoveries. However, it is very challenging to ensure the false discovery rate guarantees, because the underlying distribution is usually unknown, and the estimated non-member probabilities often exhibit interdependence. To tackle the above challenges, in this paper, we design a novel membership inference attack method, which can provide the guarantees on the false discovery rate. Additionally, we show that our method can also provide the marginal probability guarantee on labeling true non-member data as member data. Notably, our method can work as a wrapper that can be seamlessly integrated with existing MIA methods in a post-hoc manner, while also providing the FDR control. We perform the theoretical analysis for our method. Extensive experiments in various settings (e.g., the black-box setting and the lifelong learning setting) are also conducted to verify the desirable performance of our method. 
\end{abstract}

\section{Introduction}
\label{sec:intro}

Deep neural networks (DNNs) have been successfully adopted in various computer vision tasks \cite{He_2016_CVPR, zhou2022conditional,wang2023yolov7,kirillov2023segment,deitke2023objaverse,fang2023eva,zhong2023blur,cheng2024putting,kraus2023masked}. Due to the high sample complexity of such models, they require large amounts of training data. However, recent research highlights privacy risks and vulnerabilities of DNNs to membership inference attacks (MIAs) \cite{hu2022membership}. MIAs on DNNs aim to infer whether a specific data record was used to train a target model or not, thus posing severe privacy risks to individuals. For instance, if attackers infer that a clinical record has been used to train a model associated with a certain disease, they can infer that the owner of the record has that disease with high probability. A recent report \cite{tabassi2019taxonomy} published by the National Institute of Standards and Technology (NIST) specifically highlights that an MIA revealing that an individual was included in the dataset used to train the target model is a confidentiality violation.

On the other hand, beyond traditional privacy attacks, MIAs have diverse applications and play a crucial role in fields such as machine unlearning \cite{thudi2022unrolling,chen2023boundary} and lifelong learning \cite{wang2022continual}. For example, in the context of machine unlearning, MIA methods are used to evaluate the effectiveness of unlearning methods and help verify that specific samples have been successfully unlearned, thus respecting individuals’ rights to have their data removed. Note that machine unlearning refers to the process of selectively removing the influence of specific samples from trained models \cite{guo2019certified,bourtoule2021machine,warnecke2021machine,qian2023towards,zhao2023static,chen2025survey}. Additionally, in lifelong learning, where a model aims to learn continuously from new data over time while retaining previously acquired knowledge, MIAs can be used to gauge the degree of memorization for certain data. This allows us to assess how well the model retains specific information and preserves prior knowledge.

Currently, various MIA methods have been proposed. Based on the differentiation principles, existing MIA methods can be generally divided into: \emph{classifier-based}, \emph{metric-based}, \emph{likelihood ratio-based}, and \emph{quantile regression-based}. Specifically, classifier-based MIAs \cite{shokri2017membership,chen2021machine,he2022semi} usually train a binary membership classifier indicator to distinguish the behavior of training members from that of non-training members. Metric-based MIAs \cite{del2022leveraging,ko2023practical,cohen2024membership} involve defining a specific metric on model outputs to distinguish training members from non-members. Likelihood ratio-based MIAs \cite{carlini2022membership,zarifzadeh2023low} utilize parametric techniques to model the loss distributions of models that have been trained or not trained on the target test example. Quantile regression-based MIAs \cite{bertran2024scalable} utilize quantile regression on non-member distributions without training surrogate models.

Despite the significance and popularity of MIAs, existing MIA methods cannot provide guarantees on the false discovery rate (FDR), which is defined as the expected proportion of instances classified as training data (members) but are, in reality, not part of the training data (non-members) among total instances classified as training data. Traditional MIA works usually focus on empirical comparisons, and fail to provide the theoretical guarantees for member and non-member decisions. Although \cite{ye2022enhanced,bertran2024scalable} consider the ratio of non-members incorrectly identified as members, they cannot provide the guarantees on the false discovery rate. In practice, the false discovery rate provides a more precise indication of the error rate \cite{benjamini1995controlling,benjamini2010discovering,barber2015controlling,javanmard2019false}, and is crucial in settings involving simultaneous evaluations. By managing the FDR, we can mitigate the risks associated with evaluating the reliability of positive discoveries and make more informed decisions under conditions of uncertainty \cite{ma2021global,marandon2022machine,fithian2022conditional,liang2022integrative,bates2023testing}. Additionally, existing MIAs also fail to provide the marginal probability guarantee on labeling true non-member data as member data.

Our goal in this paper is to provide the guarantees on the false discovery rate for MIAs, which refers to the proportion of false discoveries among total positive discoveries in the overall testing procedure. Note that existing works on MIAs are typically framed as a hypothesis testing problem, with the alternative hypothesis asserting that the test data is from the training dataset and the null hypothesis positing it is not. However, managing the false discovery rate in the context of MIAs presents unique challenges. First, the distribution of scores for non-training data remains unknown and challenging to model accurately, which complicates the derivation of the membership indicators. Additionally, the estimated non-member probabilities usually exhibit interdependence. However, traditional multiple hypothesis testing techniques usually assume that inputs must either be independent or adhere to certain conditions of dependency. This assumption is hard to be satisfied in practice, which complicates the process of accurately controlling the FDR and makes it difficult to ensure that the proportion of false discoveries remains within acceptable limits.

To address the above challenges, in this paper, we propose \textbf{\emph{MIAFdR}}, a novel membership inference attack that can provide the false discovery rate guarantees. Specifically, in our method, given that the underlying true distribution of the member data and that of the non-member data are hard to know, we first design a novel conformity score function, which can reflect the conformity degree of test data to the non-member data. Next, based on estimated point-wise conformity scores, we present a non-member relative probability estimation strategy, which essentially reflects the likelihood of not making discoveries. We also show that based on these estimated non-member probabilities, we can provide the marginal probability guarantee on labeling true non-member data as member data. However, these generated point-wise non-member relative probabilities exhibit interdependence, making it challenging to provide the false discovery rate control. To address this, we then present an adjustment method that corrects these calculated non-member probabilities by accounting for their interdependencies and employing a weighted correction scheme. We also show that these adjusted non-member probabilities allow for controlling the false discovery rate at a predetermined significance level for prediction results. Notably, our method can be seamlessly integrated with existing MIAs to provide FDR control while preserving their attack performance. We conduct the theoretical analysis for our method. Our extensive experiments verify the effectiveness of our method. We also empirically show that our method can help data memorization-based machine learning (ML) tasks, including machine unlearning and lifelong learning.
\section{Related Work}

Membership inference attacks (MIAs) are designed to determine whether a given data sample has been used to train a particular model. The concept of MIAs is first proposed by \cite{homer2008resolving}, which aims to detect sensitive and private information leakage. Specifically, this work trains multiple shadow models to mimic the behavior of the victim model to distinguish between training samples from the training dataset and test samples. Since their inception, MIAs have gained significant attention in the research community, leading to the development of numerous MIA methods \cite{long2020pragmatic,rezaei2021difficulty,del2022leveraging,he2022semi,carlini2022membership,ko2023practical,bertran2024scalable,cohen2024membership,zhang2024generated}. Notably, numerous studies have sought to elucidate mechanisms behind MIAs, primarily attributing their operations to model memorization \cite{hayes2020remind,feldman2020neural,rezaei2021difficulty,shafran2021membership,agarwal2022estimating,carlini2023quantifying}, a phenomenon linked to overfitting. Due to memorization in DNNs, prediction confidence tends to be higher for data used for training. This difference in prediction confidence helps MIA methods to determine which image data were used for training. In \cite{yeom2018privacy}, the authors theoretically analyze the relationship between overfitting and MIAs. Therefore, beyond detecting sensitive information leakage, MIAs can also offer valuable insights into the extent of memorization in the victim model.

\begin{figure*}[t!]
% \vskip -10pt
\centering
\includegraphics[width=0.87\linewidth]{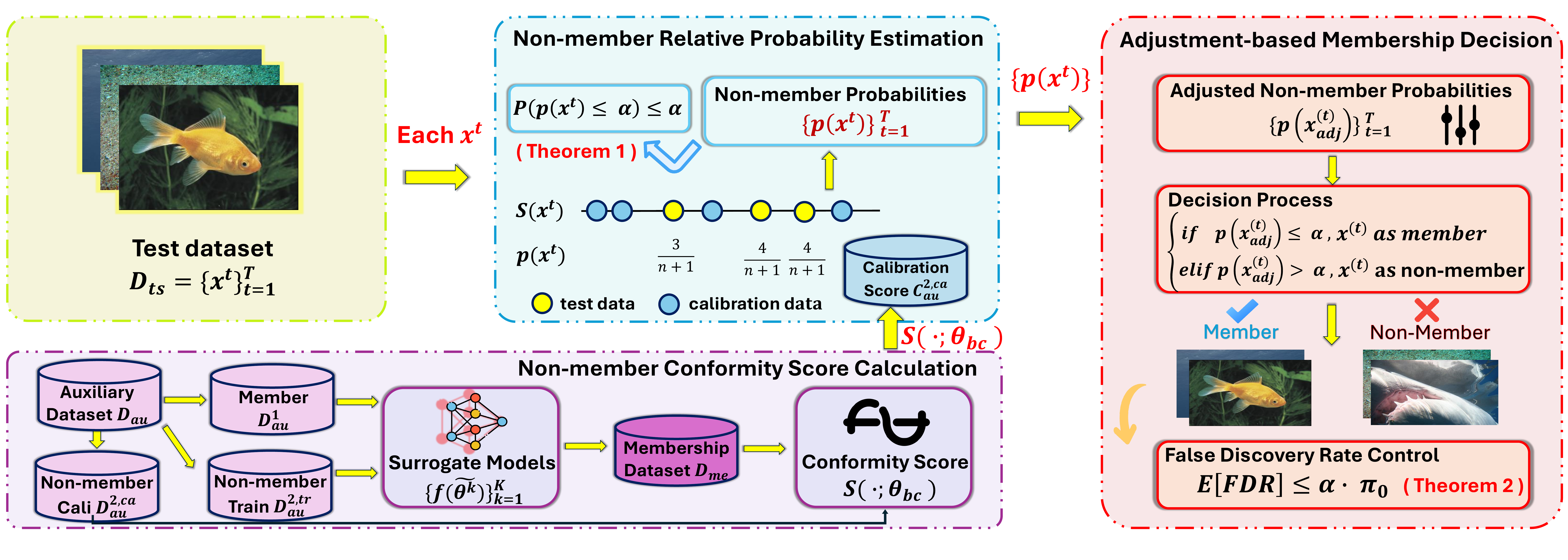}
\vskip -15pt
\caption{Overview of our proposed membership inference attacks with the false discovery rate guarantee.}
\label{fig:overview}
\vskip -10pt
\end{figure*}

Currently, MIAs have been successfully achieved in many domains and problems, including semantic image segmentation \cite{zhang2022label,he2020segmentations}, healthcare \cite{xu2023membership,gupta2021membership}, image classification \cite{rezaei2021difficulty,hu2022membership}, and recommendation systems \cite{zhang2021membership,chi2024shadow}. For example, \cite{he2020segmentations} shows that such membership inference attacks can be successfully carried out on state-of-the-art models for semantic segmentation. However, existing MIAs cannot guarantee the false discovery rate, which refers to the proportion of false discoveries among total discoveries during the attack procedure. This oversight is particularly problematic when the proportion of actual members within the test data is high. In this work, we build upon conformal inference \cite{vovk2005algorithmic,tibshirani2019conformal,gibbs2021adaptive,bastani2022practical,bates2023testing,matsumoto2023membership,kumar2023conformal,zargarbashi2024robust,qian2024towards,li2024data,wang2024bridging,chen2024modeling}, which aims to quantify the uncertainty in predictions with a specified coverage probability. However, traditional conformal inference works cannot be directly applied to provide the false discovery rate guarantees for MIAs. As aforementioned, this limitation arises because calculated non-member probabilities for test data are conditional on the shared calibration dataset and exhibit interdependence, making it challenging to provide the false discovery rate control. In contrast, our method can provide the false discovery rate guarantee, even under existing defenses against MIAs \cite{abadi2016deep,shejwalkar2021membership,niu2024survey}.

\section{Threat Model}
\label{sec:related}
% \vspace{-0.03in}
We consider a threat model that includes a model holder and an adversary. The model holder owns a well-trained DNN classifier $f(\theta^{*})$, which is trained over its training data \( D_{tr} = \{ (x_i, y_i) \}_{i=1}^{N} \) using the loss function $\mathcal{L}$ (e.g., cross-entropy loss), where $x_{i} \in \mathbb{R}^{d}$ is a $d$-dimensional feature vector and $y_{i} \in [Y]$ denotes its associated label. For a given input $x$, we can generate its softmax probability vector as
\vspace{-0.05in}
\begin{align}
  &f(x;\theta^{*}) = [f_{1}(x;\theta^{*}), f_{2}(x;\theta^{*}), \cdots, f_{Y}(x;\theta^{*})].  
\end{align}
\vskip -5pt
\noindent For the given input data \(x \), we can determine its class label as \( y(x) =\arg\max_{y \in [Y]} f_{y}(x;\theta^{*}) \), where a label \( y \in [Y]\)  with the largest probability is assigned to this given data \( x \).

The adversary aims to distinguish the training (members) and test data (non-members) of the victim model. Let \( D_{ts}=\{ x^{t}\}_{t=1}^{T} \) denote the test dataset available to the adversary. As shown in Definition \ref{def:HyTest4MIA}, for each \(x^{t} \in D_{ts}\), the adversary's goal of determining whether \( x^{t} \in D_{tr} \) or \( x^{t} \notin D_{tr} \) can be formulated as a Hypothesis Testing, where the null hypothesis (i.e., \(H_0\)) represents non-membership, while the alternative hypothesis \(H_1\) represents membership.

\begin{definition} [Hypothesis Testing for MIA]
\label{def:HyTest4MIA}
    Let $x^{t}$ denote the test data. Let \( \theta^{*}\) denote the pre-trained model trained on its private training dataset \( D_{tr} \). Then we can establish a hypothesis test where the null hypothesis posits that \(x^{t}\) was not part of the private training dataset $D_{tr}$ for the target victim model \( \theta^{*} \), as outlined below
    \vspace{-0.06in}
\begin{align}
    &H_0: x^{t} \notin D_{tr}, \text{ } \text{ } H_1: x^{t} \in D_{tr},
\end{align}
\vskip -5pt
\noindent where \( D_{tr} \) is the private training dataset for \( f(\theta^{*}) \).
\end{definition}

For the adversary's knowledge, we consider two realistic attack settings: \emph{grey-box} and \emph{black-box}. 
%\vspace{-0.028in}
\begin{itemize}
    \item In the grey-box setting, we assume that the adversary does not have access to the exact private training dataset $D_{tr}$, and has no knowledge of the victim model parameters. We consider that the adversary is aware of the knowledge of the owner’s learning algorithm and architecture.% but knows the details about the target victim model $\theta^{*}$, including its architecture and parameters. 

    %\vspace{-0.02in}
    \item In the black-box setting, we assume that the adversary does not have any prior knowledge about the private training dataset, the learning algorithm, or the target pre-trained model, including its architecture and parameters. This setting produces a realistic threat model in real-world applications, where adversaries typically operate with minimal information.
\end{itemize}
%\vspace{-0.05in}
In the above two settings, we consider a realistic adversary who has limited knowledge and does not have access to the private training dataset \( D_{tr} \). However, we allow the adversary to have access to an auxiliary dataset \( D_{au}\), distinct from the private dataset, sampled from the same distribution. 
This assumption is reasonable given the widespread availability of public data, and has been a common assumption for black-box attacks in existing literature \cite{jagielski2021subpopulation}. 
Notably, our proposed MIAFdR builds upon existing MIA methods, serving as a wrapper to ensure FDR control. Since existing MIAs~\cite{shokri2017membership,carlini2022membership,he2022semi,bertran2024scalable,zhang2024order} typically leverage on auxiliary data for training surrogate models or regression models, we similarly integrate auxiliary data into MIAFdR, aligning with the established practices.
By exploring these two settings, we aim to gain comprehensive insights into the different levels of threats posed by attackers with varying degrees of knowledge about the target model.

\section{Methodology}
\label{sec:method}

In this section, we utilize the classifier-based setting to present our method. Note that our method serves as a wrapper for existing MIAs with FDR control. \emph{Discussions on other settings are deferred to the full version of the paper}.

\textbf{Overview.} Figure~\ref{fig:overview} shows the overall framework of our proposed method, which can ensure control over the false discovery rate. Specifically, our method involves three essential components: \emph{non-member conformity score calculation}, where we design a novel conformity score function to assess the degree to which each test sample conforms to the non-member data distribution; \emph{non-member relative probability estimation}, where we utilize the previously calculated non-member conformity score to estimate the non-member relative probability for each test data, which can reflect the likelihood of this given test data being non-training data; and \emph{adjustment-based membership decision}, where we adjust the previously estimated related probabilities for these test data samples and then compare them against a predefined significance level to make membership decisions. Below, we will detail each of the three essential components.

\textbf{Non-member Conformity Score Calculation.} Note that based on Definition \ref{def:HyTest4MIA}, to determine whether \( x^{t} \) comes from \(D_{tr}\), we can establish the following hypothesis test: the null hypothesis (i.e., \( H_0 \)) posits that $x^{t}$ is not from the private training dataset $D_{tr}$ (non-members) for the victim model \( f(\theta^{*}) \), and the alternative hypothesis (i.e., \( H_1 \)) posits that it is from the private dataset. By leveraging the information of the target victim model \( f(\theta^{*}) \), we can reduce this hypothesis testing to the problem of determining whether the output (i.e., \(f(x^{t};\theta^{*}) \)) of the test data \(x^{t}\) belongs to the final softmax output distribution of the victim model \( \theta^{*}\). However, in practice, it is very difficult to obtain its underlying true final output distribution. To address this, we will estimate the empirical output distribution by considering the private training samples' final output predictions (i.e., \( \{ f(x_{i};\theta^{*}): x_{i} \in D_{tr}\}_{i=1}^{N} \)). Then we can estimate the empirical probability distribution $\hat{f}_N(\nu)$ via the average of delta functions, where \( \nu\) is any real number. Thus, we can construct the below null hypothesis testing
\vspace{-0.11in}
\begin{align}
    \label{eq:hypoDistribution}
   & \tilde{H}_0: f(x^{t};\theta^{*}) \not\sim \hat{f}_N(\nu) = \frac{1}{N} \sum_{i=1}^N \delta(\nu-f(x_{i};\theta^{*})),
\end{align}
\vskip -7pt
\noindent where \( \delta \) is the Dirac delta function, which returns \(\infty\) if the condition \( f(x_i; \theta^{*}) = \nu \) is true, and 0 otherwise.

However, in practice, the attacker usually does not have access to the private training dataset $D_{tr}$ for the victim model \( f(\theta^{*}) \), which makes it difficult to characterize the population of non-members. Therefore, it is intractable to directly adopt the hypothesis testing constructed in Eqn. (\ref{eq:hypoDistribution}).  To address this, we first train \( K \) surrogate models (denoted as \( \{ f(\tilde{\theta}^{k}) \}_{k=1}^{K} \)) and collect their predictions to estimate the empirical score distribution of non-members. Specifically, as shown in Figure~\ref{fig:overview}, to obtain \( K \) surrogate models, we first divide the auxiliary dataset \( D_{au} \) into two disjoint subsets, i.e., \( D_{au}^{1} \) and \( D_{au}^{2} \), where \( D_{au}^{1} \cap D_{au}^{2} = \emptyset \) and \( D_{au}^{1} \cup D_{au}^{2} = D_{au} \). From \( D_{au}^{1} \), we will create \( K \) subsets (i.e., \( \{ D_{au}^{1,k} \}_{k=1}^{K} \)) by sampling a fraction \(\eta\) of the data without replacement. Then, we can optimize the $k$-th surrogate model as \(\tilde{\theta}^{k} \leftarrow \arg\min_{\theta} \sum_{ (x_i,y_i) \in D_{au}^{1,k} } \mathcal{L} ((x_i, y_i);\theta)\), where \( \mathcal{L}\) is the victim model's loss assumed in the grey-box setting. Thus, we can train \( K \) surrogate models (i.e., \( \{ f(\tilde{\theta}^{k}) \}_{k=1}^{K} \)), which can approximate the behavior of the victim model \( f(\theta^{*})\).

Based on these $K$ surrogate models (i.e., \( \{ f(\tilde{\theta}^{k}) \}_{k=1}^{K} \)), for all samples within \( D_{au}^{1,k} \), we can obtain their predictions \( \mathcal{Y}_{au}^{1,k} =  \{ f( x_i;\tilde{\theta}^{k}) \}_{i=1}^{|D_{au}^{1,k}|} \), where \(\tilde{\theta}^{k}\) is the $k$-th surrogate model trained on \( D_{au}^{1,k} \). We then split the \(D_{au}^{2}\) into two disjoint sets \(D_{au}^{2,tr}\) and \(D_{au}^{2,ca}\), where \(D_{au}^{2,tr} \cap D_{au}^{2,ca} = \emptyset\), and \(D_{au}^{2,tr} \cup D_{au}^{2,ca} = D_{au}^{2}\). Similarly, for each sample \( x_i \in D_{au}^{2,tr} \) and \( x_j \in D_{au}^{2,ca} \), we can obtain \( \mathcal{Y}_{au}^{2,i} = \{ f( x_i;\tilde{\theta}^{k}) \}_{k=1}^{K} \) and \( \mathcal{Y}_{au}^{2,j} = \{ f( x_j;\tilde{\theta}^{k}) \}_{k=1}^{K} \). Thus, we have
\vspace{-0.15in}
\begin{align}
    & \mathcal{Y}_{au}^{1} = \cup_{k=1}^{K} \mathcal{Y}_{au}^{1,k}, \quad  \mathcal{Y}_{au}^{2, tr} = \cup_{i=1}^{ |D_{au}^{2,tr}| } \mathcal{Y}_{au}^{2,i},\\
    & \qquad \qquad \qquad \qquad\qquad\quad \text{ and }  \mathcal{Y}_{au}^{2, ca} = \cup_{j=1}^{ |D_{au}^{2,tr}| } \mathcal{Y}_{au}^{2,j}, \notag
\end{align}
\vskip -5pt
\noindent where \( D_{au}^{2,tr} \cup D_{au}^{2,ca} = D_{au}^{2} \subset D_{au}\). Based on this, we will construct the below membership dataset
\vspace{-0.07in}
\begin{align}
%\label{eq:MemDataset}
    & D_{me} = \{ (y_i, 0): y_i \in \mathcal{Y}_{au}^{1}  \} \cup  \{ (y_i, +1): y_i \in \mathcal{Y}_{au}^{2,tr} \}. \notag
\end{align}
\vskip -5pt
\noindent The above constructed membership dataset \( D_{me} \) can effectively capture the member distribution using the member samples labeled as $0$ and the non-member prediction using the non-member samples labeled as $+1$.

Based on the constructed membership dataset \( D_{me} =\{ z_i = (y_i,l_i) \}_{i=1}^{|\mathcal{Y}_{au}^{1}|+|\mathcal{Y}_{au}^{2,tr}|} \), where \( l_i \in \{ 0, +1\}\), we will train the following binary classifier \( f_{bc}(\theta_{bc})\) to distinguish between members and non-members
\vspace{-0.05in}
\begin{align}
\label{eq:BinaryClafer}
    & \theta_{bc} = \arg\max_{\theta} \sum_{z_i \in D_{me}} \mathcal{L}_{bc} ( z_i=(y_i, l_i) ; \theta),
\end{align}
\vskip -5pt
\noindent where \( \mathcal{L}_{bc} \) is the loss for training this classifier. Note that for \( z_i= (y_i, l_i) \in D_{me}\), it is either labeled $l_i=0$ (members) or $l_i = +1$ (non-members). Then, for the given test data \(x^{t}\), to reflect how typical it is with respect to the calibration data, we define the below non-member conformity score function 
\vspace{-0.2in}
 \begin{align}
 \label{eq:NonconfScore}
     & S(y^{t}; \theta_{bc}) = \lambda \log (\frac{f_{bc}(y^{t};\theta_{bc})}{1-f_{bc}(y^{t};\theta_{bc})}) + (1-\lambda) f_{bc}(y^{t};\theta_{bc}), 
 \end{align}
\vskip -7pt
\noindent where \( y^{t} = f(x^{t};\theta^{*})\), \( f_{bc}( \theta_{bc}) \) denotes the trained binary classifier based on Eqn. (\ref{eq:BinaryClafer}) and \(\lambda\) is a hyper-parameter to control the weight between the logit-transformed probability and raw probability. Note that for the test data \( x^{t} \), a larger non-member conformity score \( S(y^{t}; \theta_{bc}) \in \mathbb{R} \) means that it is coming from the non-member prediction; otherwise, it is more likely to be from the member distribution.

\textbf{Non-member Relative Probability Estimation.} Based on the above conformity score function \( S(\cdot;\theta_{bc}) \), for all the samples within the set \( \mathcal{Y}_{au}^{2, ca} \), we can calculate their conformity scores as \( \mathcal{C}_{au}^{2,ca} = \{ S(y_{i}; \theta_{bc}): y_{i} \in \mathcal{Y}_{au}^{2, ca} \}_{i=1}^{|\mathcal{Y}_{au}^{2, ca}|}  \). Then, we define \( P_{nm}^{*} \) as the true distribution of these conformity scores \( \mathcal{C}_{au}^{2,ca} = \{ S(y_{i}; \theta_{bc}) \}_{i=1}^{|\mathcal{Y}_{au}^{2, ca}|} \). To determine whether the given test data \(x^{t}\) is from the private dataset \( D_{tr} \), we reformulate the hypothesis test in Eqn. (\ref{eq:hypoDistribution}) into
\vspace{-0.05in}
\begin{align}
\label{eq:TrueDisBC}
\hat{H}_{0}: S(y^{t}; \theta_{bc}) \sim P_{nm}^{*}, \hat{H}_{1}: S(y^{t}; \theta_{bc}) \not\sim P_{nm}^{*},
\end{align}
\vskip -3pt
\noindent where \( P_{nm}^{*} \) is the underlying true distribution of these conformity scores \( \mathcal{C}_{au}^{2,ca} = \{ S(y_{i}; \theta_{bc}) : y_{i} \in \mathcal{Y}_{au}^{2, ca} \}_{i=1}^{|\mathcal{Y}_{au}^{2, ca}|} \).

However, the underlying true distribution \( P_{nm}^{*} \) is usually unknown, which presents significant challenges for performing the hypothesis testing procedure in Eqn. (\ref{eq:TrueDisBC}). On the other hand, directly adopting existing empirical distribution estimation methods cannot provide theoretical guarantees for the false discovery rate control, and they also usually require the assumption of underlying distributions. To address this, as demonstrated in Figure~\ref{fig:overview}, instead of estimating underlying distributions, we propose to calculate the below non-member relative probability for the test data \( x^{t}\)
\vspace{-0.066in}
\begin{align}
\label{eq:PValues}
&p(x^{t})=\frac{|\mathbb{S}^{k} \in \mathcal{C}_{au}^{2,ca}  \cup \{S(y^{t}; \theta_{bc})\}: \mathbb{S}^{k} \leq S(y^{t}; \theta_{bc}) |}{1+|\mathcal{C}_{au}^{2,ca} |},
\end{align}
\vskip -13pt
\noindent where \( \mathcal{C}_{au}^{2,ca} = \{ S(y_{i}; \theta_{bc}) \}_{i=1}^{|\mathcal{Y}_{au}^{2, ca}|} \), \(y^{t} = f(x^{t};\theta^{*})\), and \( S(y^{t}; \theta_{bc}) \) is the conformity score for the test data \(x^{t}\). The calculated non-member relative probability \(p(x^{t})\) is essentially the proportion of the calibration samples with conformity scores smaller than or equal to that of $x^{t}$. This can reflect the conformity degree of $x^{t}$ to the non-member calibration scores. Thus, we can use the calculated non-member probability \(p(x^{t})\) to assess the likelihood that $x^{t}$ was not used to train the target victim model $f(\theta^{*})$.

\begin{theorem}
\label{thm:PValueValidity}
    Let \(\mathcal{G}\) denote the sequence containing all the samples from dataset \(D_{au}^{2,ca}\) and the given test data \(x^{t}\), i.e., \(\mathcal{G} = (x^1, x^2, \ldots, x^{|D_{au}^{2,ca}|}, x^{|D_{au}^{2,ca}|+1})\), where \(|D_{au}^{2,ca}|\) is the number of samples in \(D_{au}^{2,ca}\), and \(x^{|D_{au}^{2,ca}|+1}\) is the extra term represented by \(x^{t}\). Note that \(D_{au}^{2,ca}\) is a subset of the auxiliary dataset \(D_{au}\). Assume that this sequence is exchangeable. Then, for significance level \(\alpha \in (0,1)\), we have
    \vspace{-0.07in}
    \begin{align}
    \label{eq:pvalue_validity}
    \mathcal{P}(p(x^{t}) \leq \alpha \mid x^{t} \notin D_{tr}) \leq \alpha,
    \end{align}
\vskip -4pt
\noindent where \( p(x^{t}) \) is the calculated non-member probability for test data \( x^{t}\), and $D_{tr}$ represents the private training dataset. 
\end{theorem}

In Theorem \ref{thm:PValueValidity}, we show that for a true non-member test data \( x^{t} \), the probability that its non-member relative probability \( p(x^{t}) \) is not larger than \(\alpha \) is at most \(\alpha\). For this true non-member data \( x^{t} \), it does not come from the private dataset \( D_{tr}\) (i.e., \( x^{t} \notin D_{tr}\)). This ensures that the error rate for labeling true non-member data as member data does not exceed the predefined threshold \( \alpha \), providing a guarantee on the reliability of the membership labeling process. Note that the exchangeability assumption in Theorem \ref{thm:PValueValidity} is much less restrictive than the traditional independent and identically distributed (i.i.d.) assumption \cite{de2017theory,bates2023testing,angelopoulos2021gentle}. \emph{The proof for Theorem \ref{thm:PValueValidity}, and more discussions of our method on other MIA settings are deferred to the full version of the paper}.

\textbf{Adjustment-based Membership Decision.} Next, we discuss how to provide the false discovery rate control for the test dataset \( D_{ts}=\{ x^t \}_{t=1}^{T} \), based on the above estimated non-member relative probabilities. From Definition \ref{def:FDRDefinition}, we can see that the false discovery rate is the expected proportion of false discoveries among all discoveries. However, these calculated p-values (i.e., non-member probabilities \( \{ p(x^{t})\}_{t=1}^{T} \)) for the test data \(D_{ts}\) are conditional on the calibration dataset \( D_{au}^{2,ca} \); specifically, it applies uniformly across all test data, as each calculation involves the identical calibration dataset. Consequently, the p-values obtained for the membership inference attack exhibit interdependence, which makes it challenging to provide the false discovery rate control. To control the number of false discoveries among total discoveries in the overall testing procedure, we will adjust the calculation of the original p-value in Eqn. (\ref{eq:PValues}). Specifically, we first arrange these calculated \( \{ p(x^{t})\}_{t=1}^{T} \) in ascending order, and obtain the ranked set \( \{p^{(t)}\}_{t=1}^{T} \), where $p^{(t)}$ is the non-member probability ranked at position $t$. Subsequently, for \(p^{(t)}\), as illustrated in Figure~\ref{fig:overview}, we calculate its adjusted non-member probability as
\vspace{-0.06in}
\begin{align}
\label{eq:AdjustedPValue}
& p^{(t)}_{\text {adj}}=\min \{1, \min _{m \in\{t, t+1, \ldots, n\}} \{\frac{n}{m} \cdot p^{(m)} \} \},
\end{align}
\vskip -6pt
\noindent where \( p^{(m)} \in \{p^{(t)}\}_{t=1}^{T} \) ranks at position \(m\).

\begin{definition}[False Discovery Rate (FDR)]
\label{def:FDRDefinition}
Let \(A\) denote the MIA attack algorithm, where \(A(x^t) \in\{0,1\}\), with \(0\) indicating that \(x^{t}\) is a member of the private training dataset \(D_{tr}\) of the target victim model and \(1\) indicating it is not. Let \(D_{ts}\) denote the test dataset. Then we can define the false discovery rate as \( \xi = |v_{f p}| / (|v_{f p}|+|v_{t p}|) \), where \( v_{f p}=\{x^{t} \in D_{ts} \mid A(x^{t})=0, \text{ and } x^{t} \notin D_{tr} \} \) and \( v_{tp}=\{x^{t} \in D_{ts} \mid A(x^{t})=0, \text{ and } x^{t} \in D_{tr} \} \).
\end{definition}

Based on these adjusted probabilities, for each test data \(x^{(t)} \in D_{ts}\), we can obtain its null hypothesis $H_{0, (t)}$, i.e., \( H_{0, (t)}: x^{(t)} \notin D_{tr} \). Then, for \( x^{(t)}\), we can determine 
\vspace{-0.1in}
\begin{align}
\left\{ \begin{array}{ll}
\textrm{if $p^{(t)}_{\text{adj}} \leq \alpha$} , &\textrm{$H_{0, (t)}$ does not hold,}\\ 
\textrm{if $p^{(t)}_{\text{adj}} > \alpha$}, & \textrm{ $H_{0, (t)}$ holds,} \\
\end{array} \right.
\end{align}
\vskip -7pt
\noindent where $p^{(t)}_{\text{adj}} $ is the adjusted non-member probability calculated by Eqn. (\ref{eq:AdjustedPValue}) and  $\alpha$ is a pre-defined value for determining the significance level for the multiple hypothesis testing. In the above equation, if \(p^{(t)}_{\text{adj}} \leq \alpha\), we should reject the null hypothesis, suggesting that the test data \(x^{(t)}\) does not belong to the non-members and comes from the training data \( D_{tr} \) of the victim model \(\theta^{*}\); otherwise, we should accept the null hypothesis, indicating that \( x^{(t)} \notin D_{tr} \). Based on the above, for test dataset \( D_{ts} \), we can obtain following decisions
\vspace{-0.08in}
\begin{align}
\label{eq:DecisionsR}
    & \mathcal{R}(D_{ts})= \{ t: t\in [T], \text{ and } p^{(t)}_{\text{adj}} \leq \alpha \},
\end{align}
\vskip -7pt
\noindent where \( p^{(t)}_{\text{adj}} \) is the adjusted non-member probability for the sample ranked at position \( (t) \) and \(\alpha\) is the significance level. Note that $\mathcal{R}(D_{ts})$ is the set of indices of test data within the test dataset $D_{ts}$ for which the null hypothesis is rejected. This indicates that these test samples are likely members of the private training dataset \(D_{tr}\).

\begin{theorem}
\label{thm:theorem_boundFinal} 
Let \(\mathcal{H}_0^{*}(D_{ts})=\{t: t\in [T], \text{and } H^{*}_{0, (t)} \text{is true} \}\) denote the subset of true non-members in the test data \(D_{ts}\), where \(H^{*}_{0, (t)}\) is the ground truth. Let \(\pi_0\) denote the proportion of true non-members, i.e., \(\pi_0 =  \mathcal{H}_0^{*}(D_{ts}) / T\). Then we can control the false discovery rate (FDR) at level \(\alpha \cdot \mathcal{H}_0^{*}(D_{ts}) / T\) as follows
\vspace{-0.06in}
\begin{align}
\mathbb{E}[\frac{|\mathcal{R}(D_{ts}) \cap \mathcal{H}_0^{*}(D_{ts})|}{\max \{1,|\mathcal{R}(D_{ts})|\}}]  \leq \alpha \cdot \frac{\mathcal{H}_0^{*}(D_{ts})}{T} \leq \alpha,
\end{align}
\vskip -5pt
\noindent where \( \mathcal{R}(D_{ts}) \) is the obtained positive discovery results, and $\alpha$ is the significance level in Eqn. (\ref{eq:DecisionsR}).
\end{theorem}

Theorem \ref{thm:theorem_boundFinal} states that our method allows for controlling the false discovery rate at a predetermined level for the generated positive discovery results (i.e., \( \mathcal{R}(D_{ts}) \) for test dataset  \(D_{ts}\)), thereby limiting the proportion of false discoveries among the total discoveries~\cite{bates2023testing}. \emph{The proof for Theorem \ref{thm:theorem_boundFinal}, and more discussions of our method on other MIA settings are deferred to the full version of the paper}.

\textbf{Discussion.} Note that in the above, we focus on the grey-box setting. For the threat of MIAs, we also consider the black-box setting, where attackers have no prior knowledge of the target model, including its private training data and model architecture. In this black-box setting, attackers can utilize varying architectures to train on the auxiliary dataset to approximate the target model's behavior and obtain score distributions. This strategy leverages the transferability property that arises from shared decision boundaries across different models~\cite{schwarzschild2021just, li2020towards,wang2019transferable}. In this way, attackers can effectively conduct MIAs in the black-box scenario.

\begin{figure}[htbp]
\centering
\vskip -5pt
\begin{subfigure}{0.495\linewidth}
\includegraphics[width=1\linewidth]{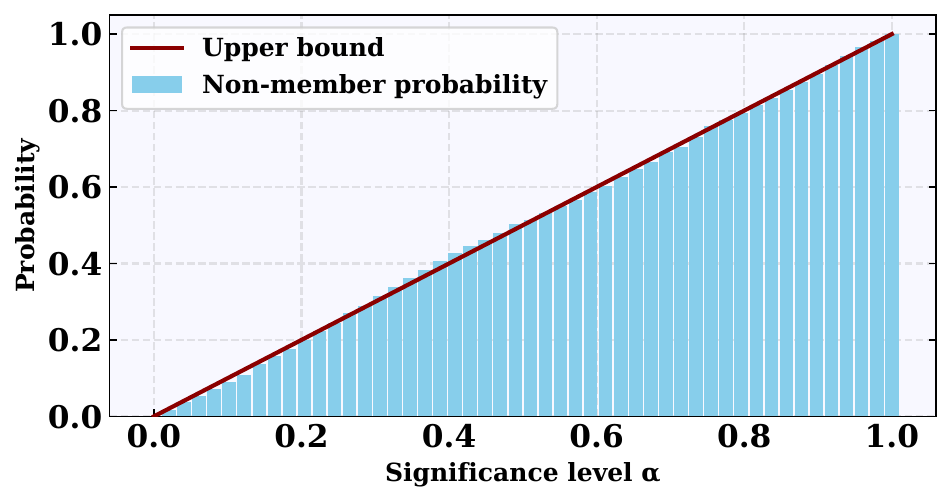}
\vskip -2pt
\caption{Statistical validity for p-value}
\label{fig:theorem_2}
\end{subfigure}
%\hfill
\begin{subfigure}{0.467\linewidth}
\includegraphics[width=1\linewidth]{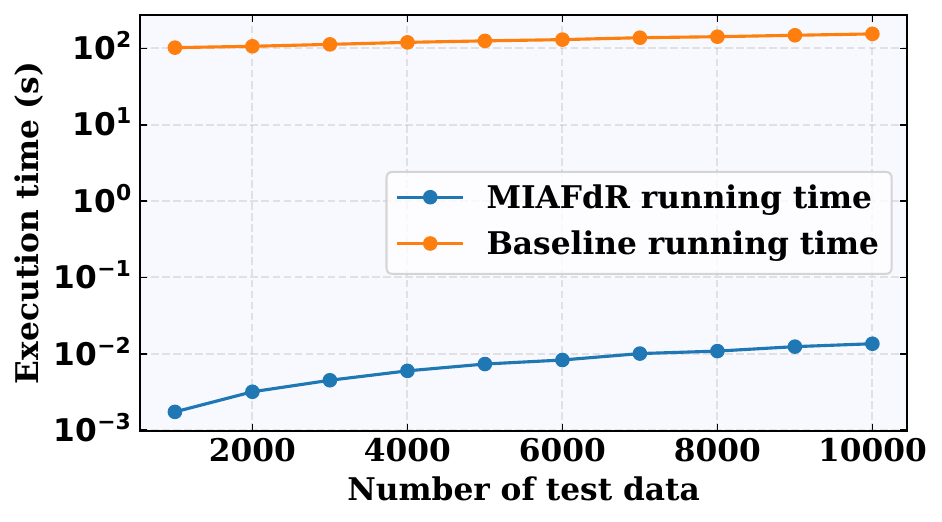}
\vskip -2pt
\caption{Running time}
\label{fig:running_time}
\end{subfigure}
\vskip -5pt
\caption{Statistical validity and running time.}
\label{fig:theorem_2_running}
\vskip -5pt
\end{figure}

\vspace{-0.05in}
\section{Experiments}
\label{sec:Exp}

In this section, we conduct extensive experiments to evaluate the effectiveness of our proposed MIAFdR. \emph{More experimental details and results (e.g., Quantile Regression-based MIAs and differential privacy-based MIA defenses) are deferred to the full version of the paper}.

\textbf{Datasets and Models.} In the experiments, we adopt the following popular benchmark image datasets: Tiny-ImageNet~\cite{deng2009imagenet}, CIFAR-100~\cite{cifar-10}, and CIFAR-10~\cite{cifar-10}. Additionally, our experimental evaluations are also conducted on various deep learning models, including ResNet-50, ResNet-18~\cite{he2016deep}, VGG-16~\cite{simonyan2014very}, MobileNetV2~\cite{sandler2018mobilenetv2}, and a multi-layer perception (MLP) network.

\textbf{Baselines.} In experiments, we compare our method with the following popular MIAs: classifier-based method such as shadow training~\cite{shokri2017membership}; metric-based methods, including Softmax~\cite{salem2018ml}, Modified Entropy (Entropy)~\cite{song2021systematic}, Loss~\cite{yeom2018privacy}, and Difficulty Calibration (Calibration)~\cite{watson2021importance}; likelihood ratio-based method represented by Likelihood Ratio Attack (LiRA)~\cite{carlini2022membership}; and quantile regression-based method like Quantile Regression Attack~\cite{bertran2024scalable}.

\begin{figure*}[htbp]
\centering
% \vskip -10pt
\begin{subfigure}{0.28\linewidth}
\includegraphics[width=0.925\linewidth]{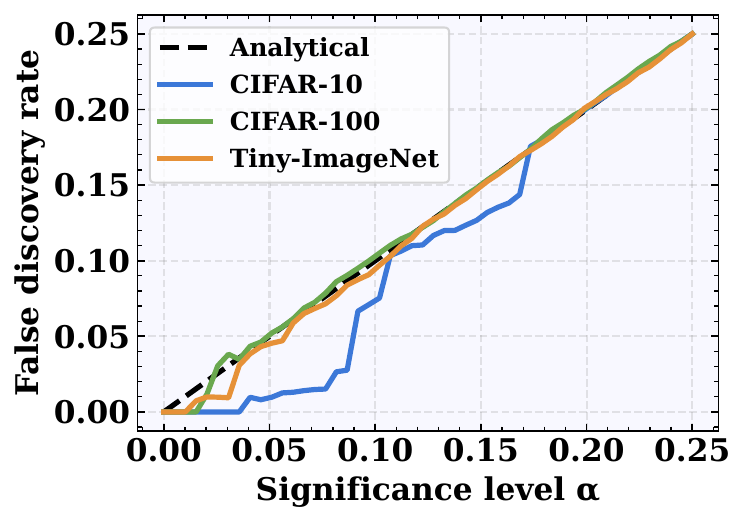}
\vskip -5pt
\caption{Classifier-based with $\pi_{0} = 0.25$}
\label{fig:classifier_conform_0.25}
\end{subfigure}
\begin{subfigure}{0.273\linewidth}
\includegraphics[width=0.9\linewidth]{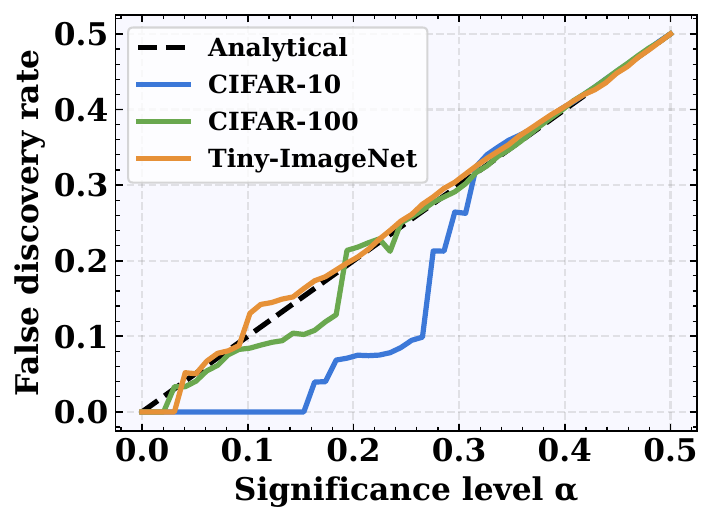}
\vskip -5pt
\caption{Classifier-based with$\pi_{0} = 0.5$}
\label{fig:classifier_conform_0.5}
\end{subfigure}
\begin{subfigure}{0.28\linewidth}
\includegraphics[width=0.9\linewidth]{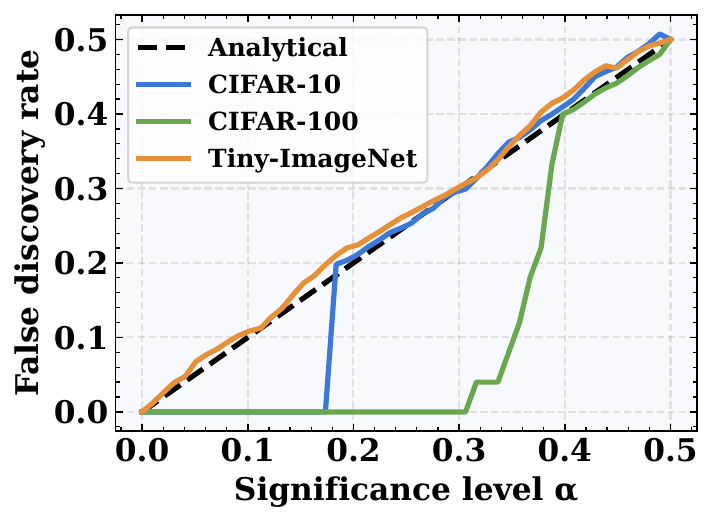}
\vskip -5pt
\caption{Metric-based with $\pi_{0} = 0.5$}
\label{fig:Softmax_bound_two_0.5_mean}
\end{subfigure}
\vskip -5pt
\caption{FDR control of classifier-based and metric-based MIAFdR.}
\label{fig:fdr_results}
\vskip -5pt
\end{figure*}

\textbf{Attack Setup.} In experiments, we split the available dataset into: a private set used for training a target model, accessible only to the model holder, and a public set employed for training a surrogate model and querying conformity scores for attackers. In likelihood ratio-based and quantile regression-based methods, the public set serves as an auxiliary set for computing conformity scores. In classified-based and metric-based methods, we allocate $40\%$ of data for calibration and $60\%$ for training the discriminative model using MLP. The evaluation is repeated 10 times, with the mean and standard deviation reported.

\vspace{-0.05in}
\subsection{Attack Performance}
\vspace{-0.05in}
First, we perform experiments to validate the Theorem \ref{thm:PValueValidity}, which establishes the validity of p-values generated by~Eqn. (\ref{eq:PValues}). We employ our proposed method with the shadow training technique~\cite{shokri2017membership} from classifier-based MIAs, utilizing the ResNet-18 model on the CIFAR-10 dataset. Initially, we partition 30\% of the auxiliary dataset \(D_{au}\) to form \(D_{au}^{1}\), allocating the remaining 70\% to \(D_{au}^{2}\). From \(D_{au}^{2}\), we then sample a fraction \(\eta = 3/7\) to obtain \(\{ D_{au}^{1,k} \}\). For each test data, we classify it as a member if its non-member probability is less than or equal to a predefined threshold $\alpha$. As shown in  Figure~\ref{fig:theorem_2}, our approach ensures that the expected error rate for false discoveries among non-training data does not exceed  \(\alpha \), thereby guaranteeing the reliability of the membership labeling process.

In addition, we aim to demonstrate the computation efficiency of our proposed method and the effectiveness of the adjusted non-member probabilities in managing the final membership decisions with respect to FDR. We follow the same experimental setup in  Figure~\ref{fig:theorem_2}. Notably, our MIAFdR serves as a wrapper for existing MIAs and maintains the same training time. In  Figure~\ref{fig:running_time}, we present an analysis of the inference runtime for the baseline method and our approach. The experimental results in  Figure~\ref{fig:running_time} show that our approach requires only a minimal additional running time compared to the baseline. For instance, with 7,000 test samples, the traditional classifier-based MIA takes 137.84 seconds, while our method adds only 0.01 seconds to the overall running time. In  Figure~\ref{fig:fdr_results}, we present the results of the FDR against varying significance level $\alpha$ for \(\pi_0 = 0.25\) and \(\pi_0=0.5\) in both classifier-based and metric-based (Softmax) settings, where \(\pi_0\) is the proportion of non-members in the test dataset. From this figure, we can see that the FDR is mostly bounded by $\alpha$, indicating that our method can effectively produce membership decisions with FDR control.

\begin{figure}[htbp]
\centering
\vskip -5pt
\begin{subfigure}{0.463\linewidth}
\includegraphics[width=1\linewidth]{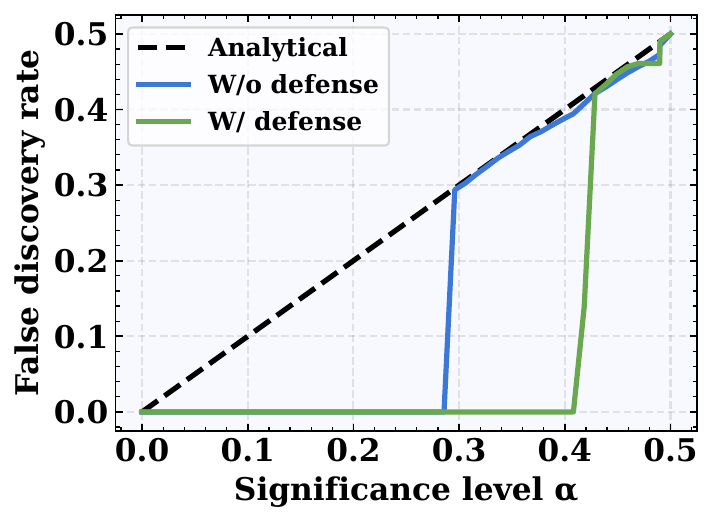}
\vskip -5pt
\caption{FDR under KD defense}
\label{fig:KD_FDR}
\end{subfigure}
%\hfill
\begin{subfigure}{0.495\linewidth}
\includegraphics[width=1\linewidth]{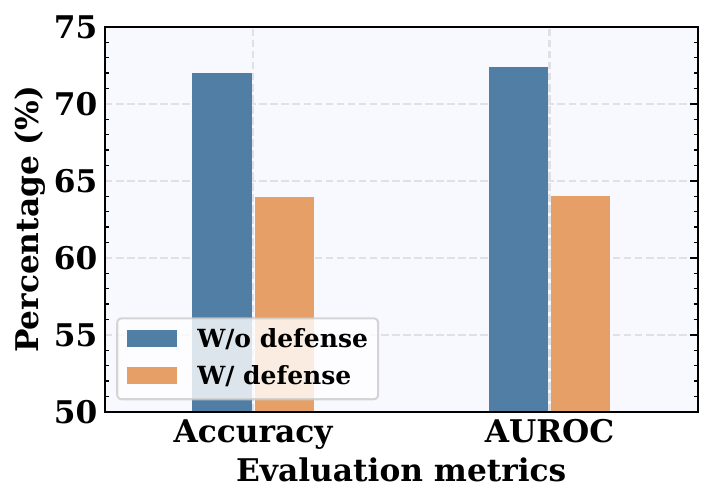}
\vskip -5pt
\caption{Metrics under KD defense}
\label{fig:KD_mertic}
\end{subfigure}
\vskip -5pt
\caption{MIAFdR against KD defense.}
\label{fig:KD_defense}
\vskip -5pt
\end{figure}

Next, we examine the effectiveness of our method in the presence of existing MIA defenses. 
We first evaluate the MIAFdR performance against the knowledge distillation-based (KD) defense. Specifically, we first train a teacher model and use its soft outputs with temperature \(T=20\) alongside the hard labels to train the student via a combined cross-entropy and KL divergence objective. The results are presented in  Figure~\ref{fig:KD_defense}.
From  Figure~\ref{fig:KD_FDR}, we can see that our method successfully maintains control over FDR even under MIA defenses. This underscores the validity of our estimated non-member relative probability and the adjustment-based membership decision, confirming their robustness even under defenses. Additionally,  Figure~\ref{fig:KD_mertic} illustrates that our method can preserve MIA prediction accuracy and AUROC under MIA defenses. More experimental results on MIAFdR against existing differential privacy-based defenses can be found in the full version of the paper.

Moreover, we investigate the attack effectiveness of MIAFdR with classifier-based MIAs, extending our analysis beyond the FDR control. Our assessment contrasts the baseline classifier-based MIA~\cite{shokri2017membership} with an enhanced version that incorporates our proposed MIAFdR across multiple datasets. The experimental results, presented in Table~\ref{tab:comp_classifier}, further demonstrate that our MIAFdR achieves comparable or even superior attack performance than the baseline classifier-based MIA in terms of attack accuracy and the AUROC score. For instance, on the CIFAR-100 dataset, our MIAFdR method attains an attack accuracy of approximately 78.2\%, outperforming the baseline accuracy of 76.8\%. These outcomes underscore the effectiveness of our classifier-based MIAFdR in accurately identifying data membership across member and non-member distributions.

\begin{table}[htbp]
\small
\centering
% \vskip -2pt
\resizebox{\columnwidth}{!}{%
\begin{tabular}{cccc} 
\toprule
Dataset & Method & Accuracy (\%) & AUROC (\%) \\ 
\midrule
\multirow{2}{*}{CIFAR-100} 
& Classifier & $76.81 \pm 1.01$ & $84.35 \pm 0.98$ \\ 
& Classifier (MIAFdR) & $\bm{78.19 \pm 0.79}$ & $\bm{84.46 \pm 0.93}$ \\ 
\midrule
\multirow{2}{*}{Tiny-ImageNet} 
& Classifier & $69.67 \pm 0.85$ & $76.99 \pm 1.63$ \\ 
& Classifier (MIAFdR) & $\bm{71.18 \pm 1.53}$ & $\bm{77.06 \pm 1.52}$ \\ 
\bottomrule
\end{tabular}
}
\vskip -6pt
\caption{Attack performance of classifier-based MIAFdR.}
\label{tab:comp_classifier}
\vskip -10pt
\end{table}

\begin{figure}[htbp]
\centering
\vskip -5pt
\begin{subfigure}{0.495\linewidth}
\includegraphics[width=1\linewidth]{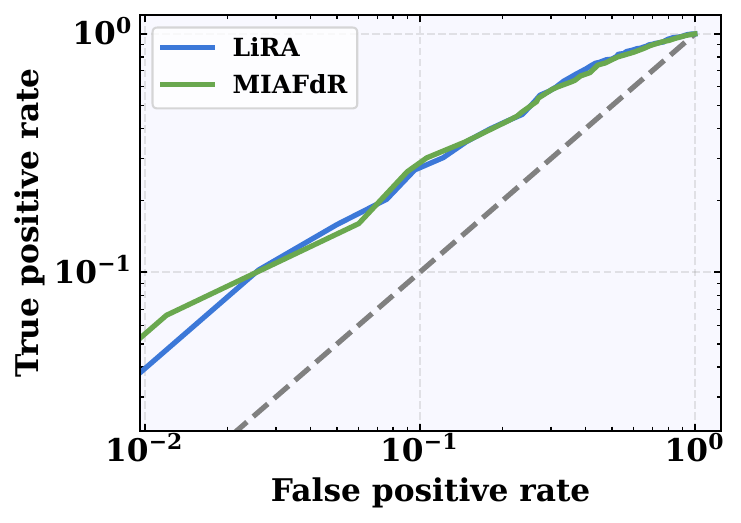}
\vskip -5pt
\caption{ROC curves on CIFAR-10}
\label{fig:first_principle_cifar10_0.5_roc}
\end{subfigure}
%\hfill
\begin{subfigure}{0.480\linewidth}
\includegraphics[width=1\linewidth]{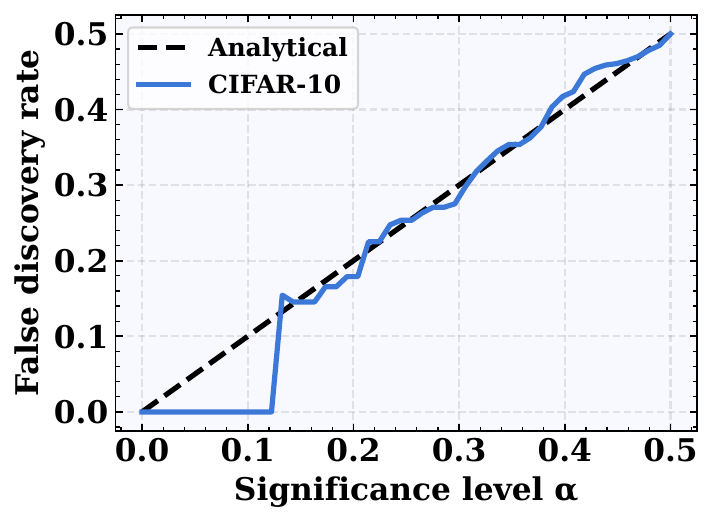}
\vskip -5pt
\caption{FDR with $\pi_{0} = 0.5$}
\label{fig:first_principle_conformal_fdr_0.5}
\end{subfigure}
\vskip -5pt
\caption{Attack performance of LiRA-based MIAFdR.}
\label{fig:first_principle}
\vskip -8pt
\end{figure}

Further, we examine the performance of our proposed MIAFdR with likelihood ratio-based MIAs. Specifically, we evaluate the performance of the original LiRA \cite{carlini2022membership} framework, which incorporates 64 shadow models, in comparison to its modified iteration, which integrates our MIAFdR approach.  Figure~\ref{fig:first_principle_cifar10_0.5_roc} depicts the ROC curves of our proposed attacks and LiRA on CIFAR-10 using log scales. As we can see, our MIAFdR with LiRA approach yields superior log-scale ROC curves, exhibiting a higher True Positive Rate (TPR) at a lower False Positive Rate (FPR). Additionally, our MIAFdR with LiRA approach effectively manages FDR, as evidenced in  Figure~\ref{fig:first_principle_conformal_fdr_0.5}. For instance, with $\pi_0=0.5$ and $\alpha=0.15$, we achieve an empirical FDR of $0.145$, which closely aligns with the analytical guarantee. Hence, our proposed attacks can be effectively integrated with likelihood ratio-based membership inference attacks to achieve both controlled FDR and good attack precision. 

\begin{figure*}[htbp]
\centering
% \vskip -5pt
\begin{subfigure}{0.28\linewidth}
\includegraphics[width=0.925\linewidth]{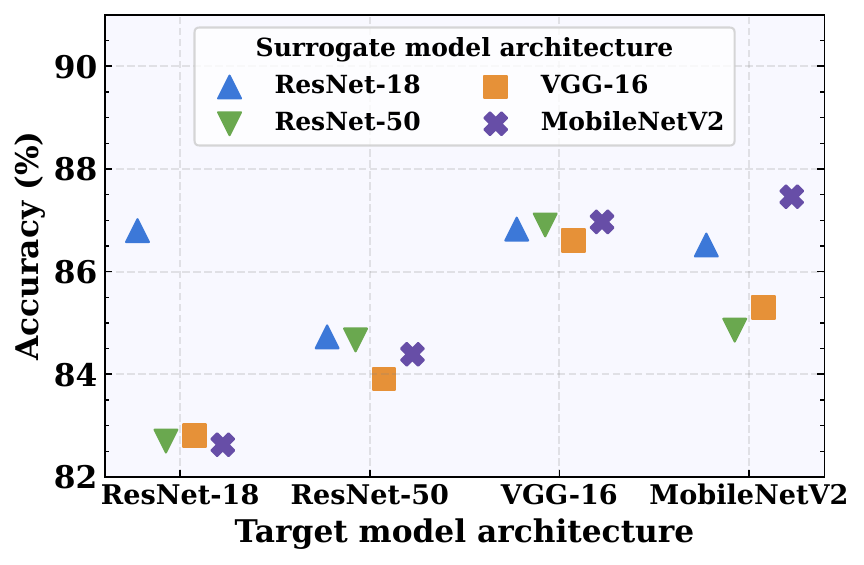}
\vskip -5pt
\caption{Black-box transferability}
\label{fig:model_arch}
\end{subfigure}
\begin{subfigure}{0.273\linewidth}
\includegraphics[width=0.9\linewidth]{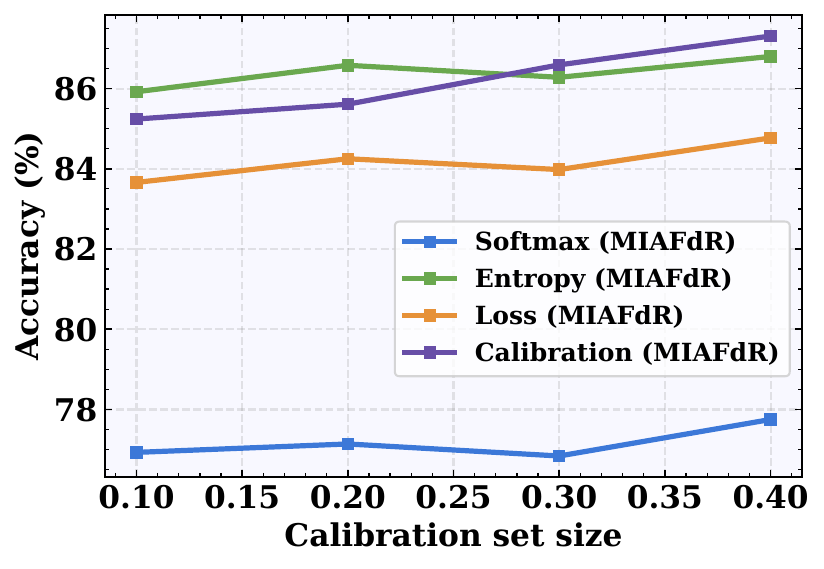}
\vskip -5pt
\caption{Impact of calibration set size}
\label{fig:cali_set_two}
\end{subfigure}
\begin{subfigure}{0.28\linewidth}
\includegraphics[width=0.9\linewidth]{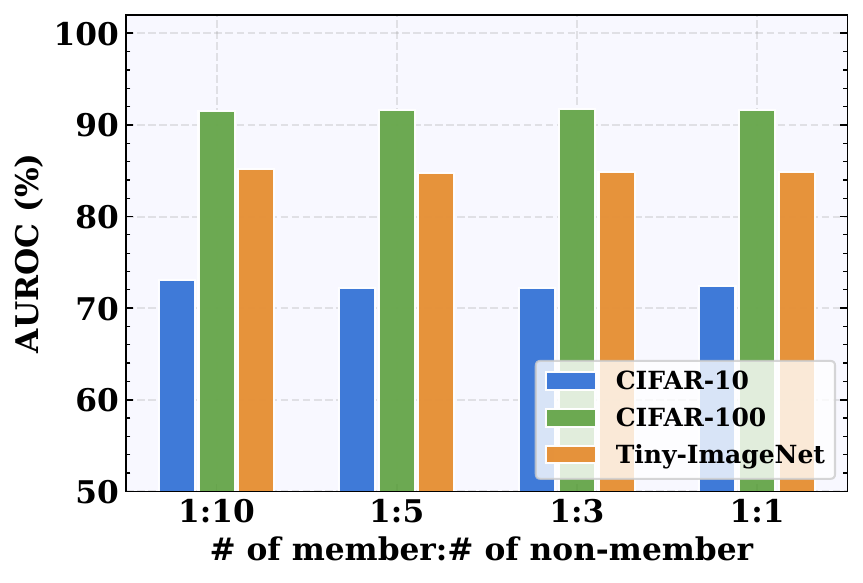}
\vskip -5pt
\caption{Impact of evaluate set}
\label{fig:eval_set_auc}
\end{subfigure}
\vskip -5pt
\caption{Black-box setting and ablation study of MIAFdR on CIFAR-10.}
\label{fig:ablation}
\vskip -10pt
\end{figure*}

At last, we conduct experiments in the black-box setting, where the attacker lacks any prior knowledge about the private training dataset or the target pre-trained model, including its architecture and parameters. Here, we train the surrogate model in MIAFdR using an architecture that differs from the target model's.  Figure~\ref{fig:model_arch} shows the attack accuracy for various model architectures on Entropy-based MIAFdR. The reported experimental results in this figure demonstrate that our MIAFdR maintains robust attack performance across various model architectures, underscoring the effectiveness of our method in the black-box setting.

\vspace{-0.05in}
\subsection{Ablation Study}
\vspace{-0.05in}
We first perform an ablation study to explore the effectiveness of MIAFdR over calibration set size and the ratio of member to non-member data. We first investigate the impact of calibration set size on metric-based MIAFdR. As shown in  Figure~\ref{fig:cali_set_two}, with a larger calibration set size, the attack accuracy tends to increase. This is because the inclusion of additional calibration data enhances the reliability of the non-member relative probability estimation, leading to more stable predictions in our MIAFdR. Next, we examine the impact of the evaluation set on MIAFdR.  Figure~\ref{fig:eval_set_auc} presents the AUROC score of MIAFdR across various ratios of member to non-member data. Remarkably, our method consistently exhibits a robust AUROC score irrespective of the ratio of member to non-member data in the evaluation set, thereby underscoring its consistent effectiveness.

\begin{figure}[htbp]
\centering
\vskip -5pt
\begin{subfigure}{0.465\linewidth}
\includegraphics[width=1\linewidth]{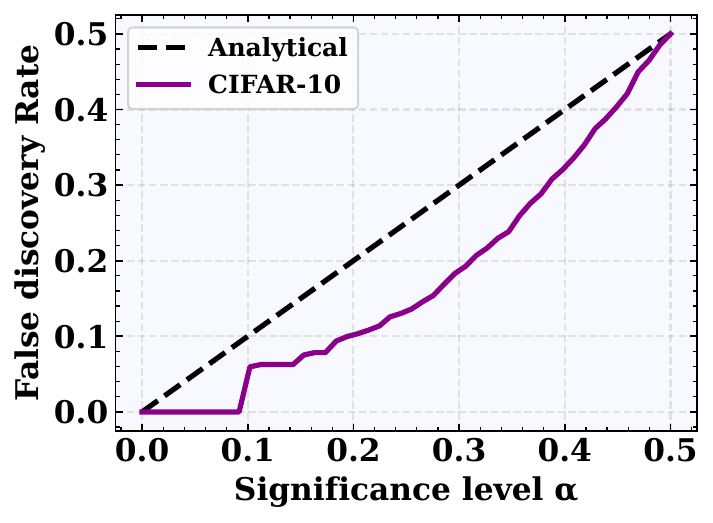}
\vskip -5pt
\caption{FDR control}
\label{fig:unlearn_FDR}
\end{subfigure}
%\hfill
\begin{subfigure}{0.495\linewidth}
\includegraphics[width=1\linewidth]{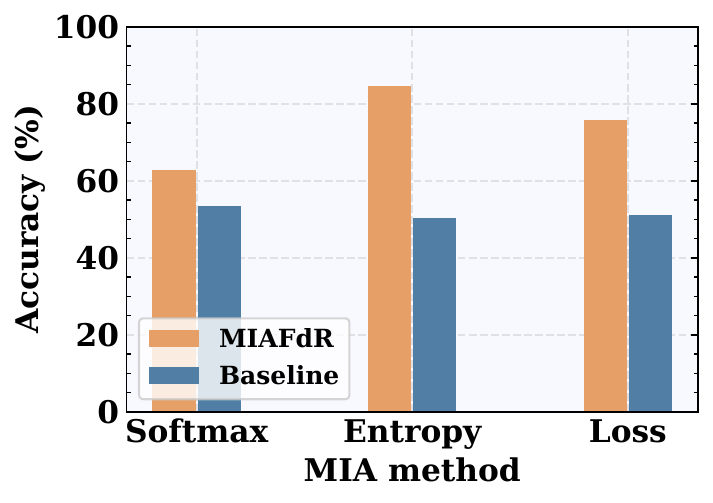}
\vskip -5pt
\caption{Accuracy}
\label{fig:unlearn_accu}
\end{subfigure}
\vskip -5pt
\caption{Machine unlearning with MIAFdR.}
\label{fig:unlearn_results}
\vskip -9pt
\end{figure}

Additionally, we conduct ablation experiments to show the impact of our MIAFdR in enhancing traditional data memorization-based ML tasks. First, in  Figure~\ref{fig:unlearn_FDR}, we report the obtained expected proportion results of instances erroneously identified as not unlearned in the context of machine unlearning. The results demonstrate that this proportion is effectively controlled across different significance levels. Here, we utilize a widely adopted popular unlearning method, i.e., SISA \cite{bourtoule2021machine}.  Figure~\ref{fig:unlearn_accu} highlights that our method significantly outperforms the baselines in accuracy.

\begin{figure}[htbp]
\centering
\vskip -5pt
\begin{subfigure}{0.495\linewidth}
\includegraphics[width=1\linewidth]{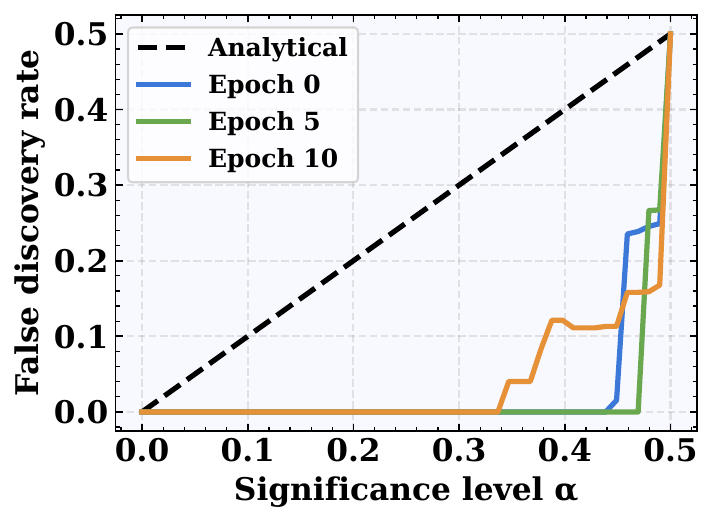}
\vskip -5pt
\caption{FDR control}
\label{fig:lifelong_FDR}
\end{subfigure}
%\hfill
\begin{subfigure}{0.488\linewidth}
\includegraphics[width=1\linewidth]{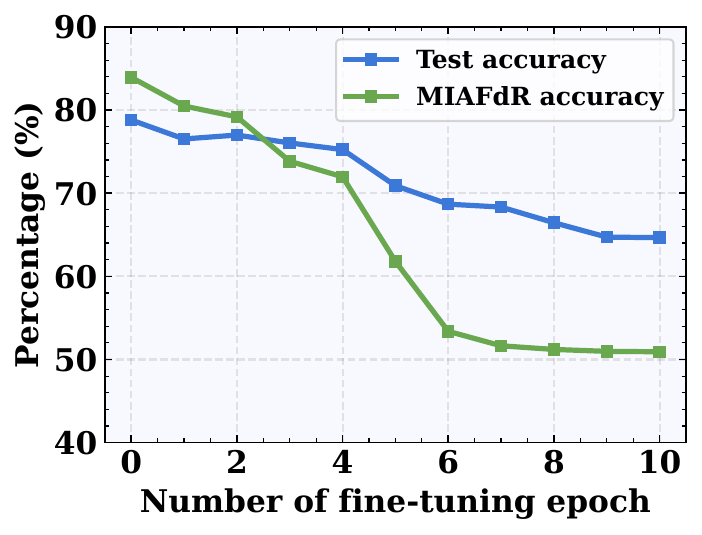}
\vskip -5pt
\caption{Accuracy}
\label{fig:lifelong_acc}
\end{subfigure}
\vskip -5pt
\caption{Lifelong learning with MIAFdR.}
\label{fig:lifelong_results}
\vskip -9pt
\end{figure}

At last, we also evaluate our method in the lifelong learning task and adopt the fine tuning-based lifelong learning method~\cite{aljundi2018memory}. For this lifelong learning setting, we report the derived experimental results in  Figure~\ref {fig:lifelong_FDR} and  Figure~\ref{fig:lifelong_acc}. Specifically,  Figure~\ref {fig:lifelong_FDR} indicates the effectiveness of our method in controlling the expected proportion of samples incorrectly reported to be memorized when they have actually been forgotten.  Figure~\ref{fig:lifelong_acc} presents the evaluation results regarding the effectiveness of lifelong learning in terms of memorizing data from previous tasks, as compared to evaluation based on accuracy. All of these experimental results verify the desired performance in traditional data memorization-based ML tasks with our proposed MIAFdR.

\vspace{-0.05in}
\section{Conclusion}
\label{sec:Conclusion}
\vspace{-0.03in}
In this paper, we design a novel membership inference attack method, which can provide the false discovery rate guarantees. Notably, our proposed MIAFdR can work as a wrapper that can be seamlessly integrated with existing MIA methods in a post-hoc manner. Specifically, in our method, given the typically unknown true distributions of member and non-member data, we first design a novel conformity score function, which can reflect the conformity degree of test data to the non-member data. Then, based on the obtained point-wise conformity scores, we develop a non-member relative probability estimation strategy to assess the likelihood of not making discoveries. Following this, we introduce a novel adjustment method that modifies the initially estimated non-member relative probabilities to ensure the false discovery rate control, effectively addressing the challenges posed by interdependent non-member relative probabilities. We conduct the theoretical analysis for our method. Extensive experiments are conducted to verify the desired performance of our method. In particular, we also empirically show that our method can help data memorization-based ML tasks, including the unlearning verification task and the lifelong learning task. 
\clearpage
\newpage

\section*{Acknowledgements}
This work is supported in part by the US National Science Foundation under grant CNS-2350332 and IIS-2442750. Any opinions, findings, and conclusions or recommendations expressed in this material are those of the author(s) and do not necessarily reflect the views of the National Science Foundation.%2350332%2442750

{
    \small
    \bibliographystyle{ieeenat_fullname}
    \bibliography{main}
}

\end{document}